\documentclass[conference]{IEEEtran}
\IEEEoverridecommandlockouts

\usepackage{cite}
\usepackage{amsmath,amssymb,amsfonts}
\usepackage{algorithm}
\usepackage{algorithmic}
\usepackage{graphicx}
\usepackage{textcomp}
\usepackage{xcolor}
\usepackage{hyperref}
\hypersetup{hidelinks=true}
\usepackage{makecell}
\usepackage{tabularx}
\usepackage{booktabs}
\usepackage{array}
\def\BibTeX{{\rm B\kern-.05em{\sc i\kern-.025em b}\kern-.08em
    T\kern-.1667em\lower.7ex\hbox{E}\kern-.125emX}}

\begin{document}

\title{A Contrastive Variational AutoEncoder for NSCLC Survival Prediction with Missing Modalities}

\author{%
\IEEEauthorblockN{Michele Zanitti\IEEEauthorrefmark{1}, Vanja Miskovic\IEEEauthorrefmark{2}, Francesco Trov\`o\IEEEauthorrefmark{2}, Alessandra Laura Giulia Pedrocchi\IEEEauthorrefmark{2},\\ Ming Shen\IEEEauthorrefmark{1}, Yan Kyaw Tun\IEEEauthorrefmark{1}, Arsela Prelaj\IEEEauthorrefmark{3}, and Sokol Kosta\IEEEauthorrefmark{1}}
\IEEEauthorblockA{\IEEEauthorrefmark{1}Department of Electronic Systems, Aalborg University, Copenhagen, Denmark}
\IEEEauthorblockA{\IEEEauthorrefmark{2}Department of Electronics, Information and Bioengineering, Politecnico di Milano, Milan, Italy}
\IEEEauthorblockA{\IEEEauthorrefmark{3}Department of Medical Oncology, Istituto Nazionale dei Tumori, Milan, Italy}
}

\maketitle

\begin{abstract}
Predicting survival outcomes for non-small cell lung cancer (NSCLC) patients is challenging due to the different individual prognostic features. This task can benefit from the integration of whole-slide images, bulk transcriptomics, and DNA methylation, which offer complementary views of the patient's condition at diagnosis. However, real-world clinical datasets are often incomplete, with entire modalities missing for a significant fraction of patients. State-of-the-art models rely on available data to create patient-level representations or use generative models to infer missing modalities, but they lack robustness in cases of severe missingness. We propose a Multimodal Contrastive Variational AutoEncoder (MCVAE) to address this issue: modality-specific variational encoders capture the uncertainty in each data source, and a fusion bottleneck with learned gating mechanisms is introduced to normalize the contributions from present modalities. We propose a multi-task objective that combines survival loss and reconstruction loss to regularize patient representations, along with a cross-modal contrastive loss that enforces cross-modal alignment in the latent space. During training, we apply stochastic modality masking to improve the robustness to arbitrary missingness patterns. Extensive evaluations on the TCGA-LUAD ($n=475$) and TCGA-LUSC ($n=446$) datasets demonstrate the efficacy of our approach in predicting disease-specific survival (DSS) and its robustness to severe missingness scenarios compared to two state-of-the-art models. Finally, we bring some clarifications on multimodal integration by testing our model on all subsets of modalities, finding that integration is not always beneficial to the task.
\end{abstract}


\section{Introduction}
\label{sec:introduction}

Applications of artificial intelligence in oncology have massively accelerated in recent years. This progress is largely driven by an increased research interest in multimodal learning, which aims to simultaneously integrate diverse data inputs, representing a transition from conventional unimodal models that analyze single data types in isolation~\cite{Truhn2024}. In oncology, such systems leverage complementary information spanning histopathology, radiology, and molecular profiles, often surpassing single-modality baselines in diagnostic, prognostic, and treatment selection tasks~\cite{Lipkova2022, Boehm2021, Truhn2024, Steyaert2023, Vale-Silva2021, Vanguri2022, MSabah2021}. Whole-slide images provide the histological features necessary to assess tumor morphology, cellular structure, and visible gene mutations. Instead, omics data provide molecular markers to discover more fine-grained cancer processes. DNA methylation instead encodes critical epigenetic mechanisms driving cancer initiation and progression~\cite{Lakshminarasimhan2016}. Each modality offers partial understanding of the tumor status, making their integration essential for accurate prognosis and biomarker discovery~\cite{Steyaert2023}.

However, real-world patient records are frequently incomplete, posing a fundamental challenge to the applicability of multimodal systems. Patient records may lack entire modalities due to numerous factors, including patient-specific treatment paths, data acquisition methods, cost, or privacy constraints~\cite{Huang2020,Flores2023,Wu2024}. A recent pan-cancer study found that only $35.9\%$ of patients had complete multimodal data for time-to-next-treatment analysis~\cite{Keyl2025}, highlighting the pervasiveness of this issue. In this light, restricting the analysis to patients with all modalities may reduce sample sizes dramatically and introduce selection bias by excluding patients whose missingness patterns correlate with survival outcomes.

Current approaches to handling missing modalities can be categorized into two main strategies, each with significant limitations~\cite{Wu2024}. Representation methods train models to use only available modalities, but may learn biased representations that favor frequently present data sources, potentially overlooking other modalities due to the modality collapse phenomenon~\cite{Vale-Silva2021, Wumm2024, Choi2019}. Imputation-based approaches attempt to reconstruct missing modalities from available data but risk introducing systematic biases when cross-modal relationships are non-linear or differ across patient subgroups~\cite{Flores2023, Zhang2022}.

To address these limitations, we propose Multimodal Contrastive Variational AutoEncoder (MCVAE), a representation learning framework designed for incomplete multimodal data. MCVAE learns modality-agnostic latent representations that adapt to arbitrary patterns of modality availability through two key components: a variational framework to model the uncertainty in each modality's contribution, and an adaptive fusion mechanism to weight available modalities based on the learned importance and availability. Our multi-task objective combines survival prediction with reconstruction regularization and a contrastive learning objective to ensure cross-modal alignment in the learned representations. We train the MCVAE with a masked modality regime, where the model learns to reconstruct the missing modalities when they are available. Our model can be seen as a hybrid approach between modality dropout and imputation methods, which neither simply relies on available modalities nor attempts to impute missing ones.

We validate MCVAE on two non-small cell lung cancer (NSCLC) cohorts from The Cancer Genome Atlas (TCGA): lung adenocarcinoma (LUAD) and lung squamous cell carcinoma (LUSC).\footnote{\url{https://portal.gdc.cancer.gov/}.} Our experiments demonstrate that MCVAE achieves state-of-the-art performance in disease-specific survival prediction, but also maintains stability as missingness severity increases, with minimal performance degradation when more than $80\%$ of modalities are missing.

Our contributions are summarized as follows:
\begin{itemize}
    \item We introduce MCVAE, a variational framework that learns robust modality-agnostic representations to account for scenarios with arbitrary patterns of missing data.
    \item We propose an adaptive fusion mechanism with availability-aware gating that normalizes contributions from present modalities.
    \item Our multi-task learning objective combines survival prediction with regularization via reconstruction and cross-modal alignment.
    \item Through extensive experiments on TCGA-LUAD and TCGA-LUSC cohorts, we demonstrate that MCVAE achieves equal or higher performance against two state-of-the-art methods~\cite{Wumm2024, Zhang2022} for survival prediction, remaining stable as missingness rates increase.
    \item We provide comprehensive modality ablation studies revealing that multimodal integration does not universally improve performance, with implications for future work.
\end{itemize}

\section{Related Work}
\subsection{Unimodal learning}
Training models on unimodal data has been the standard approach in cancer research. 
In histopathology, foundation models such as GigaPath\cite{Xu2024} and UNI\cite{Chen2024} demonstrate state-of-the-art performance in tumor subtyping and survival prediction from WSIs. Similarly, in radiology, RadImageNet\cite{Mei2022} and CT-FM\cite{Pai2025} are two foundation models pre-trained on over one million annotated CT, MRI, and ultrasound scans to support large-scale transfer learning across downstream classification and segmentation tasks. For molecular profiles, graph-based models such as XGRAD\cite{Guan2024} leverage gene co-expression networks for cancer subtype prediction from RNA sequencing data. In structured clinical data, SCORPIO~\cite{Yoo2025}, an ensemble of machine learning models, is trained to predict patient response to immune checkpoint inhibitors without requiring extensive diagnostic information, achieving reasonable performance despite the task difficulty. While the achievements of unimodal models are impressive, they are fundamentally limited by the information content of their single data type. This recognition has driven the field towards multimodal integration: in oncology, multimodal learning aims to integrate diverse data types—histopathology, omics, and clinical features to produce comprehensive patient representations.

\subsection{Multimodal Integration Strategies}
Early fusion approaches concatenate features from different modalities before processing, as demonstrated in MultiSurv~\cite{Vale-Silva2021}, which combines transcriptomics and clinical data for survival prediction. Chen et al.~developed MCAT~\cite{Chen2021}, an attention-based approach that combines WSI features and genomics to predict survival outcomes in five TCGA cohorts. Graph-based methods have also gained recognition; MUSE~\cite{Wumm2024} models patient-modality relationships as a bipartite graph, using graph neural networks to aggregate information from available modalities.
Despite these advances, most multimodal methods assume complete data availability during training or testing. However, obtaining complete cases is unrealistic since data acquisition methods differ for individual patients~\cite{Lipkova2022, Zhang2022}, resulting in sparse modality representations. This gap has motivated the development of methods specifically designed for incomplete multimodal data.

\subsection{Handling Missing Modalities}
The challenge of missing modalities has inspired several solutions, which can be broadly categorized into representation-based and imputation-based approaches~\cite{Wu2024}.

Representation-based methods train models to function with arbitrary subsets of modalities. MultiSurv~\cite{Vale-Silva2021} uses modality masking during training to handle missingness, but its concatenation-based fusion weights all available modalities equally, regardless of their informativeness. Similarly, EmbraceNet~\cite{Choi2019} introduces a stochastic sampling from available modality embeddings to mitigate reliance on single sources. MUSE~\cite{Wumm2024} extends this concept by incorporating a self-supervised objective that reconstructs modality embeddings using local similarity patterns. Despite their flexibility, these methods can be biased toward frequently available modalities and cannot explicitly model the uncertainty introduced by missing data.

Imputation-based approaches explicitly reconstruct missing modalities from available ones. SMIL~\cite{Ma2021} is a probabilistic model that learns to reconstruct missing modalities via class-specific priors learned during pretraining on complete samples, which can misrepresent the latent distribution when applied to patients with different missingness patterns. M3Care~\cite{Zhang2022} constructs patient similarity graphs for each modality and imputes missing representations in latent space based on similar patients with complete data, but the quality of imputation degrades significantly when available modalities have few similar complete cases. While these methods attempt to recover information from missing modalities, reconstructed modalities may not faithfully capture the true data distribution, leading to representational bias that favors common phenotypes.

To address the limitations of both types of methods, we propose a new model based on variational inference~\cite{Kingma2013} for the available modalities. 

\section{Methods}
\label{sec:methods}

\begin{figure*}[!t]
\centering
\includegraphics[width=\textwidth]{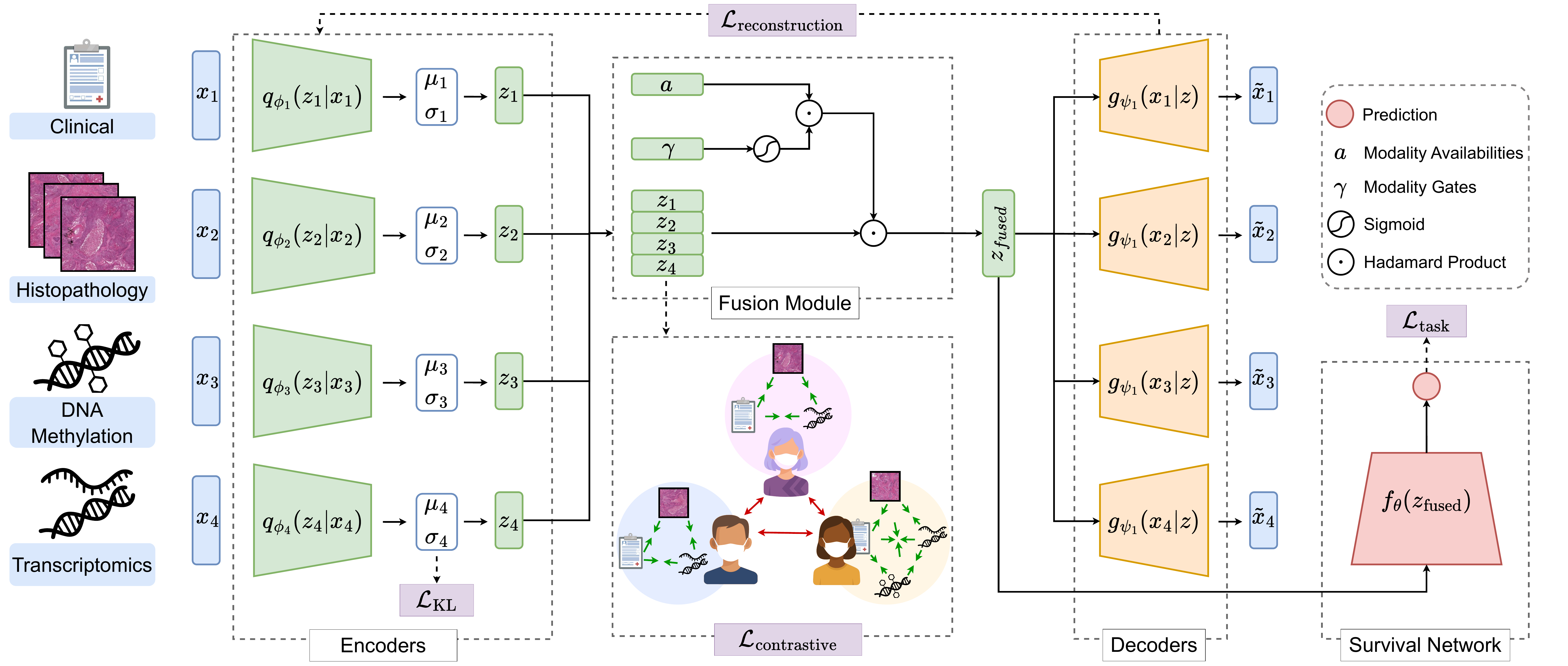}
\caption{Overview of the proposed MCVAE architecture. From the left: each modality is encoded through a modality-specific variational encoder, producing latent variables $z_k$. The fusion module combines available modality embeddings using availability-aware gating and a fusion network to obtain a shared latent representation $z_{\text{fused}}$. This representation is used for modality reconstruction via decoders and for survival prediction via the survival head. Training optimizes a composite objective that includes task loss (Cox partial likelihood), reconstruction loss, KL divergence, and InfoNCE contrastive loss. The latter pulls together embeddings from different modalities of the same patient (green arrows) while pushing apart embeddings from different patients (red arrows). Loss components are highlighted in purple.}
\label{fig:model_architecture}
\end{figure*}

\subsection{Problem Formulation}
In this paper, we aim to predict survival outcomes in the presence of incomplete multimodal data for NSCLC patients. For the $i$-th patient, we model the hazard function $h_i(t)$ which represents the instantaneous risk of death at time $t$, given their available clinical information.

Let $\mathcal{D} = \{(X_i, a_i, t_i, \delta_i)\}_{i=1}^N$ denote our dataset of $N$ patients, where: $X_i = \{x_i^{(1)}, x_i^{(2)}, x_i^{(3)}, x_i^{(4)}\}$ represents the multimodal data with $x_i^{(1)} \in \mathbb{R}^{d_c}$ (clinical features), $x_i^{(2)} \in \mathbb{R}^{d_t}$ (transcriptomics), $x_i^{(3)} \in \mathbb{R}^{d_w}$ (WSI features), and $x_i^{(4)} \in \mathbb{R}^{d_m}$ (methylation). $a_i \in \{0,1\}^4$ is the availability mask where $a_i^{(k)} = 1$ indicates modality $k$ is available. $t_i \in \mathbb{R}^+$ is the observed time (either time to death or last follow-up) and $\delta_i \in \{0,1\}$ is the event indicator ($\delta_i = 1$ for death observed, $\delta_i = 0$ for censored).

Our objective is to learn a function $f_\theta: (X_i, a_i) \rightarrow \mathbb{R}$ that predicts the log-hazard ratio accounting for arbitrary patterns of missing data.
Under the Cox proportional hazards framework~\cite{Cox1972}, we model:
\begin{equation}
\label{eq:cox-hazard}
h_i(t_i | X_i, a_i) = h_0(t_i) \exp(f_\theta(X_i, a_i)),
\end{equation}
where $h_0(t_i)$ is the baseline hazard and $f_\theta(\cdot,\cdot)$ is our learned function. The survival probability is then:
\begin{equation}
\label{eq:survival-prob}
S_i(t_i | X_i, a_i) = \exp\!\big(-H_i(t_i | X_i, a_i)\big),
\end{equation}
where $H_i(t_i | X_i, a_i) := -\int_0^{t{_i}} h_i(u | X_i, a_i)\,\mathrm{d}u$ is the cumulative hazard function.

\subsection{Multimodal Contrastive Variational Autoencoder (MCVAE)}
We propose a multimodal fusion framework designed to handle incomplete data. Figure~\ref{fig:model_architecture} illustrates the overall architecture. Unlike existing models that either discard information through dropout (e.g., Embracenet) or introduce bias through imputation, MCVAE learns robust shared representations by explicitly modeling uncertainty and adapting its fusion strategy based on arbitrary patterns of modality availability.

\subsubsection{Variational Encoding with Uncertainty Quantification}
For each available modality $k$, we learn a latent representation $z^{(k)} \in \mathbb{R}^{d_{out}}$ using a variational encoder~\cite{Kingma2013}:
\begin{equation}
\label{eq:variational-encoder}
q_{\phi_k}(z^{(k)} | x^{(k)}) = \mathcal{N}(\mu^{(k)}, \text{diag}(\sigma^{2(k)})),
\end{equation}
where $\mu^{(k)} = f_{\mu}^{(k)}(x^{(k)})\in \mathbb{R}^{d_{out}}$ and $\log(\sigma^{(k)})^{2} = f_{\sigma}^{(k)}(x^{(k)})\in \mathbb{R}^{d_{out}}$ are outputs of neural networks parameterized by $\phi_k$. Under this formulation, larger posterior variances ($(\sigma^{(k)})^{2}\gg 0$) indicate lower confidence in the modality-$k$ representation. We sample from these distributions using the standard reparameterization trick: $z^{(k)} = \mu^{(k)} + \sigma^{(k)} \odot \epsilon$, with Gaussian noise $\epsilon \sim \mathcal{N}(0, I)$.\footnote{We use the notation from~\cite{Kingma2013}, where the symbols $\odot$ and $\sim$ signify element-wise
product and sampling from a distribution, respectively.} For missing modalities, we set $z^{(k)} = 0$ and handle them through our fusion mechanism.

\subsubsection{Adaptive Fusion with Availability-Aware Gating}
Modalities for patient $i$ are aggregated to produce a latent embedding $z_{(i,\text{fused})}$ as follows:

\begin{equation}
\label{eq:adaptive-fusion}
    z_{(i,\text{fused})} = \mathbf{h}\left( \frac{1}{\sum_{k=1}^K a^{(k)}} \sum_{k=1}^K a^{(k)} \sigma(\gamma_k) z^{(k)} \right),
\end{equation}
where $\gamma_k$ are learnable modality-specific gate parameters that capture the relative importance of each modality, scaled into the range $(0, 1)$ via $\sigma(\cdot)$, $\mathbf{h}(\cdot)$ is a fusion network that transforms modality representations once aggregated, and the normalization term $C := \frac{1}{\sum_{k=1}^K a^{(k)}}$ ensures consistent scaling regardless of how many modalities are present. This design addresses key limitations of existing methods, e.g., unlike binary presence indicators as in MUSE~\cite{Wumm2024}, our gating mechanism provides graduated importance scores and avoids explicit reconstruction of missing modalities as in M3Care~\cite{Zhang2022} or SMIL~\cite{Ma2021}.

\subsubsection{Multi-Task Learning Objective}
MCVAE is trained as a multi-task learning objective with four loss components.

\textbf{Survival Prediction Loss}: For the primary task of survival prediction, we minimize the negative log partial likelihood of the Cox model, formally defined as follows:
\begin{equation}
\label{eq:task-loss}
\mathcal{L}_{\text{task}} = -\sum_{i: \delta_i=1} \left[ f_{\theta}(z_{(i,\text{fused})}) - \log \sum_{j \in \mathcal{R}_i} \exp(f_{\theta}(z_{(j,\text{fused})}) \right],
\end{equation}
where the function $f_{\theta}$ operates on the fused representation $z_{(\cdot,\text{fused})}$, and $\mathcal{R}_i = \{j : t_j \geq t_i\}$ is the set of patients still at risk at the event time $t_i$.

\textbf{Reconstruction Regularization}: To preserve modality-specific information in the shared representation, we attach a decoder $g_{\psi_k}: \mathbb{R}^{d_z} \to \mathbb{R}^{d_k}$ to produce the reconstruction $\tilde{x}_k$ for each observed modality $k$ and penalize the reconstruction error, formally defined as follows:
\begin{equation}
\label{eq:recon-loss}
\mathcal{L}_{\text{recon}} = \sum_{k=1}^K a^{(k)} \|x^{(k)} - g_{\psi_k}(z_{(i,\text{fused})})\|_2^2.
\end{equation} 
Let us remark that, in this architecture, we do not reconstruct missing modalities ($a^{(k)} = 0$), so gradients are not propagated through their decoders. This term acts as a consistency regularizer under modality dropout, encouraging the fused latent to retain information predictive of each available modality.

\textbf{KL Divergence with Modality-Specific Weighting}: We regularize the latent distributions with learnable modality-specific weights as follows:
\begin{equation}
\label{eq:kl-loss}
\mathcal{L}_{\text{KL}} = \sum_{k=1}^K a^{(k)} \cdot w_k \cdot \text{KL}(q_{\phi_k}(z^{(k)}|x^{(k)}) \| p(z)),
\end{equation}
where $w_k := \text{softmax}(\mathbf{w}_{\text{KL}})_k$ allows the model to adjust regularization strength based on each modality's reliability. Here, we selected $p(z) = \mathcal{N}(0, I)$ as the prior distribution over latent variables.

\textbf{Contrastive Cross-Modal Alignment}: To ensure consistency across modalities from the same patient, we employ the InfoNCE loss~\cite{vandenOordDeepMind2018}, formally defined as follows:
\begin{equation}
\label{eq:contrastive-loss}
\mathcal{L}_{\text{contrast}}
= - \sum_{i}\sum_{k \in \mathcal{A}_i} \sum_{\substack{l \in \mathcal{A}_i \\ l \neq k}}
\log \frac{\exp\!\big(s_{(i,k),(i,l)}/\tau\big)}
{\sum_{\substack{j=1,\dots,N \\ m \in \mathcal{A}_j \\ (j,m)\neq(i,k)}}
\exp\!\big(s_{(i,k),(j,m)}/\tau\big)} ,
\end{equation}
where $\mathcal{A}_i = \{k : a_i^{(k)} = 1\}$ denotes the set of available modalities for patient $i$. We define $s_{(i,k),(j,m)} = \mathrm{sim}(z_{(i,k)},z_{(j,m)})$ as the cosine similarity between latent representations of modality $k$ from patient $i$ and modality $m$ from patient $j$, and $\tau$ is a temperature parameter. The numerator compares pairs of modalities $(k,l)$ from the same patient $i$ (positives), while the denominator includes all other patient-modality pairs $(j,m)$ in the batch. This loss encourages representations from different modalities of the same patient to be similar while pushing apart representations from different patients. 

The multi-task training objective combines these loss terms:
\begin{equation}
\label{eq:composite-loss}
\mathcal{L} = \mathcal{L}_{\text{task}} + \lambda_1 \mathcal{L}_{\text{recon}} + \lambda_2 \beta(t) \mathcal{L}_{\text{KL}} + \lambda_3 \mathcal{L}_{\text{contrast}},
\end{equation}
where $\mathcal{L}_{\text{task}}$ ensures accurate survival prediction, $\mathcal{L}_{\text{recon}}$ preserves modality-specific information, $\mathcal{L}_{\text{KL}}$ regularizes the latent space to prevent posterior collapse, and $\mathcal{L}_{\text{contrast}}$ aligns patient representations across modalities. The KL term includes an annealing factor $\beta(t)$ that gradually increases from 0 to 1 during training to prevent posterior collapse.

We adopt the approach from~\cite{Kendall2018}, where each loss term in Eq.~\ref{eq:composite-loss} is weighted using dynamic coefficients instead of linear $\lambda_i$ weights to account for task-related uncertainties:
\begin{equation}
\label{eq:adaptive-weighting}
\mathcal{L}_{\text{total}} = \sum_{i} \left( \frac{1}{2\sigma_i^2} \mathcal{L}_i + \log \sigma_i\right),
\end{equation}
where $\sigma_i$ are learnable parameters that balance different objectives based on their uncertainty.

\subsection{Training Procedure}
During training, we apply stochastic modality dropout to improve robustness. More specifically, for each patient $i$ and for modality $k$, we sample from a Bernoulli random variable $D \sim \text{Bernoulli}(1 - p_{\text{drop}})$ with dropout probability $p_{\text{drop}}$:
\begin{equation}
\label{eq:modality-dropout}
\tilde{a}^{(k)} = \begin{cases}
1 & \text{if } k = 1 \text{ (clinical)} \\
a^{(k)} \cdot D & \text{otherwise}
\end{cases}.
\end{equation}
This ensures that each non-clinical modality has a probability $p_{\text{drop}}$ of being masked during training, even when it is originally available. We always retain clinical features ($k = 1$), as they are most reliably available in clinical practice and serve as a stable anchor for learning. The pseudocode for MCVAE's training and inference steps is provided in Algorithm~\ref{alg:mcvae}.

\begin{algorithm}[!t]
\caption{Training and Inference for MCVAE}
\label{alg:mcvae}
\begin{algorithmic}[1]
\STATE \textbf{// Training}
\REQUIRE Training data $\mathcal{D}$, dropout rate $p_{\text{drop}}$
\FOR{each epoch $t$}
    \FOR{each batch}
        \STATE Sample batch $(X, a, t, \delta)$ from $\mathcal{D}$
        \STATE Apply modality dropout by Eq.~\ref{eq:modality-dropout}, keep clinical
        \STATE Encode modalities to $\{\mu^{(k)}, \sigma^{(k)}\}$ by Eq.~\ref{eq:variational-encoder}
        \STATE Sample latents $z^{(k)}$ via reparameterization
        \STATE Fuse representations to $z_{\text{fused}}$ by Eq.~\ref{eq:adaptive-fusion}
        \STATE Compute losses by Eq.~\ref{eq:task-loss},~\ref{eq:recon-loss},~\ref{eq:kl-loss},~\ref{eq:contrastive-loss}
        \STATE Weight losses by Eq.~\ref{eq:composite-loss},~\ref{eq:adaptive-weighting}
        \STATE Update model parameters 
    \ENDFOR
\ENDFOR
\STATE \textbf{// Inference}
\REQUIRE Test data $\mathcal{D}_{\text{test}}$
\FOR{each batch}
    \STATE Sample batch $(X, a)$ from $\mathcal{D}_{\text{test}}$
    \STATE Encode available modalities by Eq.~\ref{eq:variational-encoder}
    \STATE Fuse representations by Eq.~\ref{eq:adaptive-fusion}
    \STATE Predict survival via trained model
\ENDFOR
\end{algorithmic}
\end{algorithm}

\section{Experimental Setup}
\label{sec:exp}

\subsection{Datasets and Preprocessing}
We evaluated our method on two public datasets from The Cancer Genome Atlas (TCGA): lung adenocarcinoma (LUAD\, $n=475$) and lung squamous cell carcinoma (LUSC, $n=446$). Following recent work~\cite{Jaume2023}, we selected disease-specific survival (DSS) as our primary clinical endpoint, as broader measures like overall survival (OS) capture death events from all causes and lack cancer-specificity. We remark that for LUAD and LUSC, DSS is computed through approximation, as mentioned in~\cite{Liu2018}, since the cause of death was not documented for individual patients. However, it remains a clinically relevant endpoint, whereas it may be unreliable for other TCGA cancer types with insufficient events or shorter follow-up periods.

\textbf{Clinical metadata collection}: Clinical covariates and survival labels were obtained from the UCSC Xena database~\cite{Goldman2020}, including patient demographics (e.g., age, sex, ethnicity), tumor staging, smoking habits, and basic treatment information.

\textbf{Histopathology collection}: Primary diagnostic WSIs were downloaded from the Imaging Data Commons (IDC) portal~\cite{Fedorov2021}. We processed all images using the Trident package~\cite{Zhang2025} at $20\times$ magnification, extracting $256 \times 256$ patches without overlap. To extract patch features, we selected the UNI foundation model~\cite{Chen2024}. Unlike other foundation models, UNI is not pre-trained on TCGA, making it a suitable choice in light of concerns about data contamination~\cite {Campanella2025}. Finally, we aggregated patch-level features into slide-level representations via mean pooling.

\textbf{Genomic data collection}: We obtained the gene expression counts from the Xena database and reproduced the gene set enrichment analysis from~\cite{Jaume2023} to extract relevant pathways. We used gene set collections from the Molecular Signature Database (MSigDB) Hallmarks and Reactome. Of the extracted pathways, we retained those with at least $90\%$ of gene coverage, resulting in $50$ (Hallmark) and $1783$ (Reactome) pathways.

\textbf{DNA Methylation collection}: We acquired Illumina HumanMethylation450 BeadChip data via the UCSC Xena database, containing beta values for approximately $450,000$ CpG sites per sample. To the raw beta values, we applied the following pipeline and extracted complementary features: we aggregated probes into gene-level methylation scores for promoter regions with $\geq 90\%$ probe coverage (adapted from MethylMix~\cite{Cedoz2018}), computed CpG density-based metrics on the bimodal distribution~\cite{Bibikova2011}, and derived cancer-specific methylation signatures~\cite{Weisenberger2006}.

Feature extraction for clinical, genomic, and methylation data was performed after train-test splitting to prevent data leakage. Clinical features were normalized using robust scaling. Transcriptomic pathway scores were computed as the mean expression of pathway genes. Methylation features were extracted using the combined approach described above. The resulting multimodal feature sets $\mathcal{D}$, including aggregated histopathology features, were then used for downstream survival analysis.

Natural missingness in the datasets reflects real-world scenarios: approximately $15\%$ of patients lack methylation data, $8\%$ lack transcriptomics, and $3\%$ lack WSI features. Clinical data is available for all patients.

\subsection{Training Protocol and Implementation}
We implemented $5$-fold cross-validation following recent recommendations~\cite{Gross2024}, with stratification by survival outcome (death within median survival time) to ensure balanced representation. For each fold, we allocated $64\%$ of the patients for training, $16\%$ for validation, and $20\%$ for testing. We repeat the entire process with $3$ different random seeds, reporting the mean and standard deviation across all $15$ runs.

For training our models, we used the AdamW optimizer with weight decay regularization.
We employ Harrell's concordance index (C-index) as the primary metric for evaluating survival prediction. A C-index of $0.5$ indicates random performance, while $1.0$ indicates perfect ranking.
We used the validation sets to early stop the model on the best C-index achieved, with a patience of $20$ epochs. The model checkpoint with the largest validation C-index is used for final evaluation.
All experiments were conducted on a single A10 GPU with 24GB of video memory, as we found this was sufficient for extracting WSI features with the UNI model and training all models.

\subsection{State-of-the-art Methods}
We compared MCVAE against two state-of-the-art methods for each strategy of handling missing modalities: MUSE (representation-based)~\cite{Wumm2024} and M3Care (imputation-based)~\cite{Zhang2022}. For fair comparison, all models were trained with identical modality-specific encoder architectures and underwent systematic hyperparameter optimization using the Optuna framework~\cite{Akiba2019}, with $50$ iterations of Bayesian search on the validation set. All reported results use the optimal hyperparameters identified for each model. Further details on the implementations and final hyperparameters are listed in Appendices~\ref{appendix_implementation} and~\ref{appendix_hp}.

\subsection{Experiment Design}
We designed four complementary experiments to evaluate our approach.

\subsubsection{Survival Prediction} 
We compare MCVAE to the state-of-the-art models on the primary task. Statistical significance between models is assessed using a Friedman test on the fold-level C-index values. When the Friedman test indicates significant differences ($p < 0.05$), we apply the Nemenyi post-hoc test to identify which specific model pairs differ significantly while controlling for multiple comparisons.

\subsubsection{Modality Combination Analysis} 
To understand the contribution of individual modalities and their interactions to survival prediction performance, we conduct an ablation study evaluating all possible combinations that include clinical data, ensuring the same number of patients across all combinations. Let us define the following modalities: C = clinical, T = transcriptomics, W = whole-slide images, and M = methylation. We use the same hyperparameters identified during the survival prediction analysis and evaluate the following configurations: C (unimodal clinical), C+T, C+W, C+M, C+T+W, C+T+M, C+W+M, and C+T+W+M, where the + sign indicates that the model used multiple modalities as input.

\subsubsection{Modality Dropout Analysis} 
To test the hypothesis that stochastic masking of entire modalities regularizes the shared encoder and yields a modality-agnostic latent suitable for survival prediction, we train models with dropout probabilities $p_{\text{drop}} \in \{0.0, 0.1, 0.3, 0.5, 0.7, 0.9\}$. At each step, a random subset of modalities is dropped, except for clinical data. We evaluate the robustness under natural missingness, setting the performance on complete-case data as the baseline (i.e., no dropout is applied), and testing the robustness–accuracy trade-off as $p_{\text{drop}}$ increases. This design isolates the effect of synthetic missingness during training on representation quality and downstream survival prediction, allowing identification of the dropout regime that best balances generalization to missing data with performance when all modalities are present.

\subsubsection{Progressive Modality Missingness Analysis} While the dropout analysis serves the purpose of finding an optimal masking probability for training a model, this analysis allows us to systematically evaluate model degradation under increasingly severe modality missingness. In this case, we set the dropout probability during training to $0.3$. We controlled random missingness rates using the same dropout probability as in Equation~\eqref{eq:modality-dropout}, masking $\{10\%, 30\%, 50\%, 70\%, 90\%\}$ of the available modalities, while always retaining the clinical features. This protocol enables assessment of the absolute performance at each missingness level, as well as the stability of each studied model as $m$ increases. At $m = 90\%$, the evaluation approximates a worst-case regime in which, for most samples, clinical data are the only modality available.

\begin{figure}[th!]
\centerline{\includegraphics[width=\columnwidth]{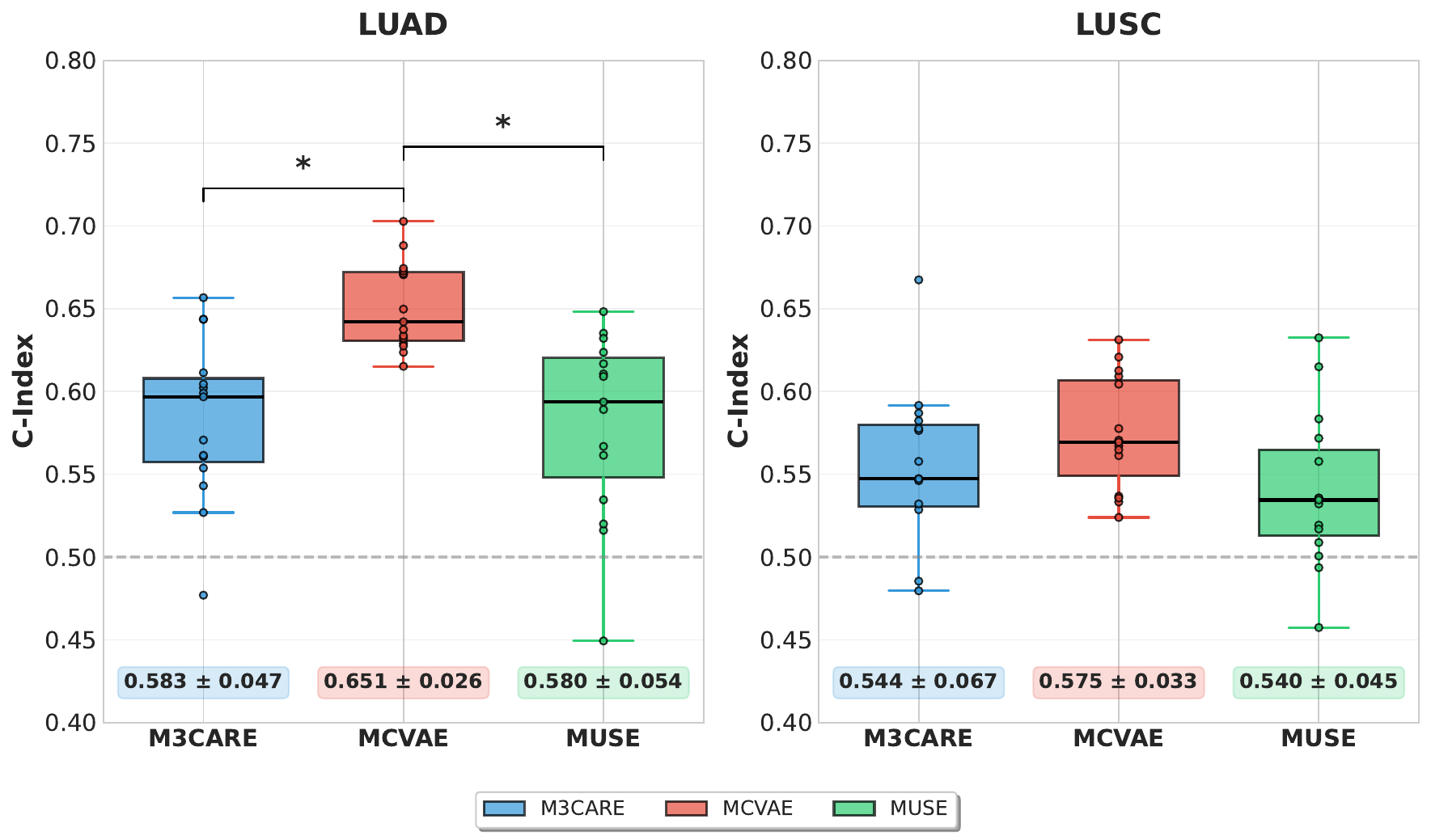}}
\caption{Results on survival analysis. C-index and individual results on each fold are reported. The asterisks (*) indicate a significant increase in performance between the two models.}
\label{surv_comparison}
\end{figure}

\section{Simulation Results}
\label{sec:results}

\subsection{Survival Prediction Performance} 
Figure~\ref{surv_comparison} presents the cross-validation results for DSS prediction across both NSCLC cohorts. We set the modality dropout probability to $0.3$ for all models. The Friedman test indicates significant differences among models for both cohorts (LUAD: $p < 0.001$, LUSC: $p < 0.001$), with post-hoc Nemenyi tests revealing specific pairwise differences.
On the LUAD cohort, MCVAE achieves the highest mean C-index of $0.651 \pm 0.026$, representing a substantial improvement over MUSE and M3Care. On the LUSC cohort, performance differences are less pronounced but still meaningful. MCVAE achieves $0.575 \pm 0.033$, with M3Care showing higher mean performance than MUSE, which exhibits a substantially higher variance.
We observe that MCVAE demonstrates more stable performance across folds, as indicated by the lower standard deviation, suggesting that our fusion approach yields more consistent predictions. Second, MUSE and M3Care exhibit significant performance variability across the two cohorts, suggesting that modeling patient similarities with graph convolutions (as in M3Care and MUSE) yields conflicting results, possibly due to unique patient-level features (e.g., genotype). Finally, we observe a performance gap between the two cohorts, as all models have a lower achieved C-index on LUSC, which is in line with previous studies~\cite{Gross2024, Jaume2023}.

\subsection{Modality Combination Analysis} 
Tables~\ref{luad_ablation} and~\ref{lusc_ablation} report the average and standard deviation of C-index obtained with repeated cross-validation on LUAD and LUSC, respectively. The best model for each combination is highlighted in bold. Since the goal of this analysis is to compare the performance obtained with all modalities, we first compare the results of the main analysis with the unimodal baseline.

\begin{table}[th!]
\centering
\caption{Results by modality combinations on TCGA LUAD.}
\label{luad_ablation}
\begin{tabular}{lccc}
\toprule
Modalities & M3Care & MCVAE (ours) & MUSE \\
\midrule
C &  0.607 $\pm$ 0.034 & \textbf{0.669 $\pm$ 0.013} & 0.634 $\pm$ 0.066 \\
\midrule
C + T  & 0.611 $\pm$ 0.051 & \textbf{0.663 $\pm$ 0.022} & 0.552 $\pm$ 0.092 \\
C + W  & 0.586 $\pm$ 0.043 & \textbf{0.652 $\pm$ 0.028} & 0.597 $\pm$ 0.071 \\
C + M  & 0.595 $\pm$ 0.072 & \textbf{0.671 $\pm$ 0.037} & 0.578 $\pm$ 0.087 \\
\midrule
C + T + W  & 0.566 $\pm$ 0.052 & \textbf{0.647 $\pm$ 0.021} & 0.608 $\pm$ 0.046 \\
C + T + M  & 0.626 $\pm$ 0.050 & \textbf{0.668 $\pm$ 0.027} & 0.534 $\pm$ 0.100 \\
C + W + M  & 0.577 $\pm$ 0.062 & \textbf{0.655 $\pm$ 0.027} & 0.603 $\pm$ 0.054 \\
\midrule
All  & 0.587 $\pm$ 0.047 & \textbf{0.651 $\pm$ 0.026} & 0.580 $\pm$ 0.054 \\
\bottomrule
\end{tabular}
\end{table}

\begin{table}[th!]
\centering
\caption{Results by modality combinations on TCGA LUSC.}
\label{lusc_ablation}
\begin{tabular}{lccc}
\toprule
Modalities  & M3Care & MCVAE (ours) & MUSE \\
\midrule
C  & \textbf{0.564 $\pm$ 0.033} & 0.557 $\pm$ 0.031 & 0.549 $\pm$ 0.092 \\
\midrule
C + T  & \textbf{0.583 $\pm$ 0.064} & 0.566 $\pm$ 0.033 & 0.553 $\pm$ 0.058 \\
C + W  & \textbf{0.548 $\pm$ 0.058} & 0.541 $\pm$ 0.056 & 0.536 $\pm$ 0.053 \\
C + M  & 0.562 $\pm$ 0.050 & \textbf{0.570 $\pm$ 0.041} & 0.551 $\pm$ 0.071 \\
\midrule
C + T + W  & 0.542 $\pm$ 0.062 & \textbf{0.555 $\pm$ 0.043} & 0.520 $\pm$ 0.057 \\
C + T + M  & 0.532 $\pm$ 0.065 & 0.594 $\pm$ 0.046 & \textbf{0.607 $\pm$ 0.060} \\
C + W + M  & 0.556 $\pm$ 0.064 & \textbf{0.558 $\pm$ 0.040} & 0.552 $\pm$ 0.060 \\
\midrule
All   & 0.544 $\pm$ 0.067 & \textbf{0.575 $\pm$ 0.033} & 0.540 $\pm$ 0.045 \\
\bottomrule
\end{tabular}
\end{table}

With only clinical covariates, all models already achieve a strong predictive performance that often surpasses multimFor LUAD, all models have unimodal C-indices ranging from $0.607$ (M3Care) $0.607$ (M3Care) up to $0.669$ (MCVAE). On LUSC, clinical features yield C-indices around $0.55$-$0.56$ across all models.
With the addition of more modalities, we observe a slight performance degradation: all models struggle to achieve better results than the unimodal baseline on LUAD, with average C-index differences of $-1.8\%$ (MCVAE), $-2\%$ (M3Care), and $-5.4\%$ (MUSE). On LUSC, this pattern is less pronounced but still present for MUSE. We performed a one-sided Wilcoxon signed-rank test to verify this assumption, revealing significant degradation ($p < 0.05$, Holm-corrected) for MUSE with C+T ($-12.97\%$), M3Care with C+T+W ($-6.81\%$), and MCVAE with C+T+W ($-3.34\%$) on LUAD.
MCVAE demonstrates the most robust multi-modal integration when all four modalities are considered and maintains a competitive performance with M3Care in both cohorts. M3Care performs well with bimodal combinations on LUSC, particularly C+T and C+W, but its performance degrades with three or more modalities on LUAD. MUSE is particularly sensitive to multi-modal integration on LUAD ($-15.81\%$ for C+T+M), but interestingly achieves the best performance on LUSC with C+T+M.
The two cohorts exhibit different responses to multi-modal integration: while on LUAD, all models perform worse with more modalities integrated, only MCVAE shows improved results on LUSC. We noticed that when WSI is included, models tend to perform worse, indicating a weaker signal from that modality. On the other hand, the C+T+M combination shows mixed results, causing the worst drop for MUSE on LUAD but the best improvement on LUSC.

\begin{figure}[!ht]
\centerline{\includegraphics[width=\columnwidth]{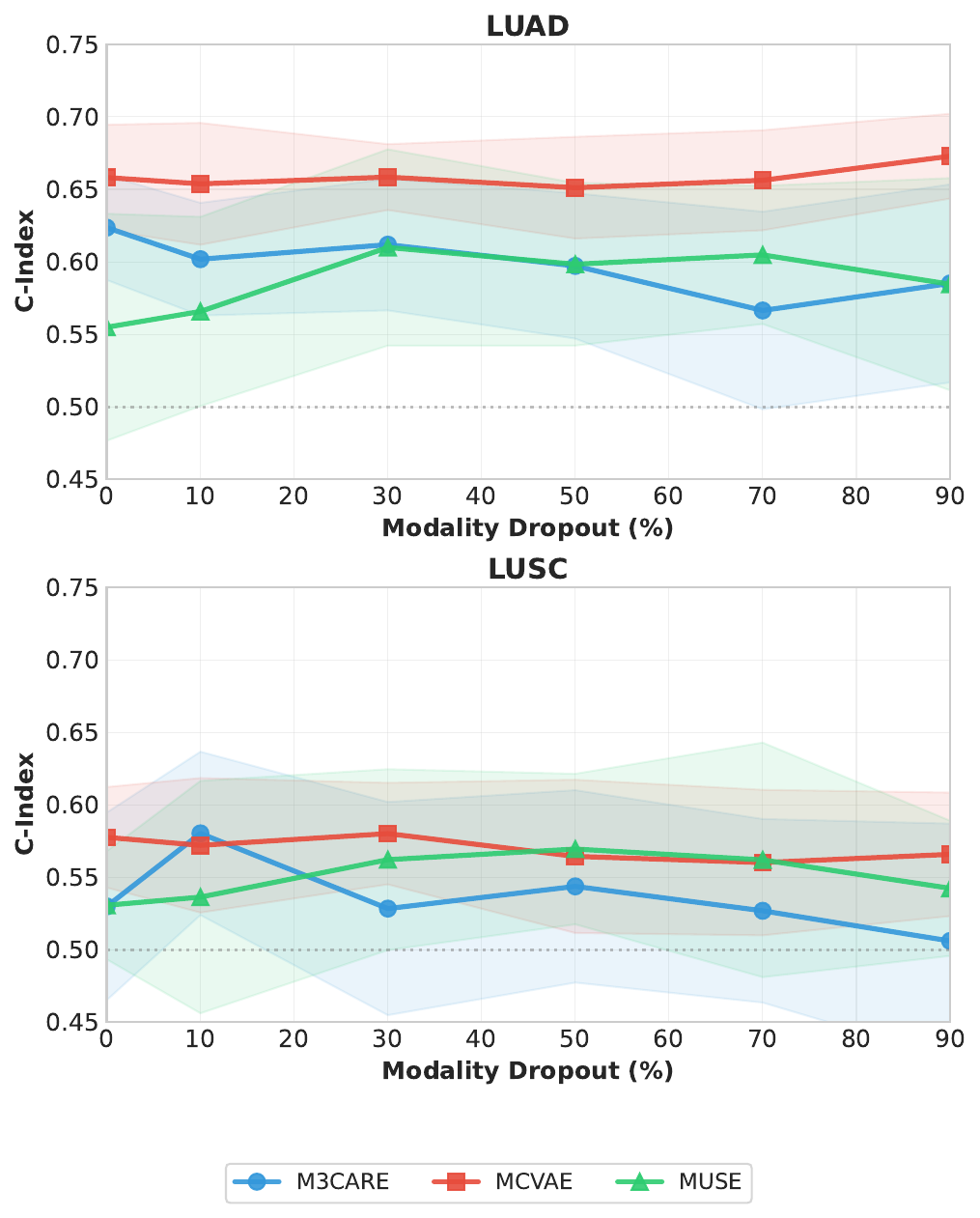}}
\caption{Results on survival analysis for LUAD and LUSC cohorts for different modality dropout rates $p_{drop}$. Average C-index (bold markers) and standard deviations (shaded area) are reported.}
\label{mod_drop}
\end{figure}

\begin{figure}[!ht]
\centerline{\includegraphics[width=\columnwidth]{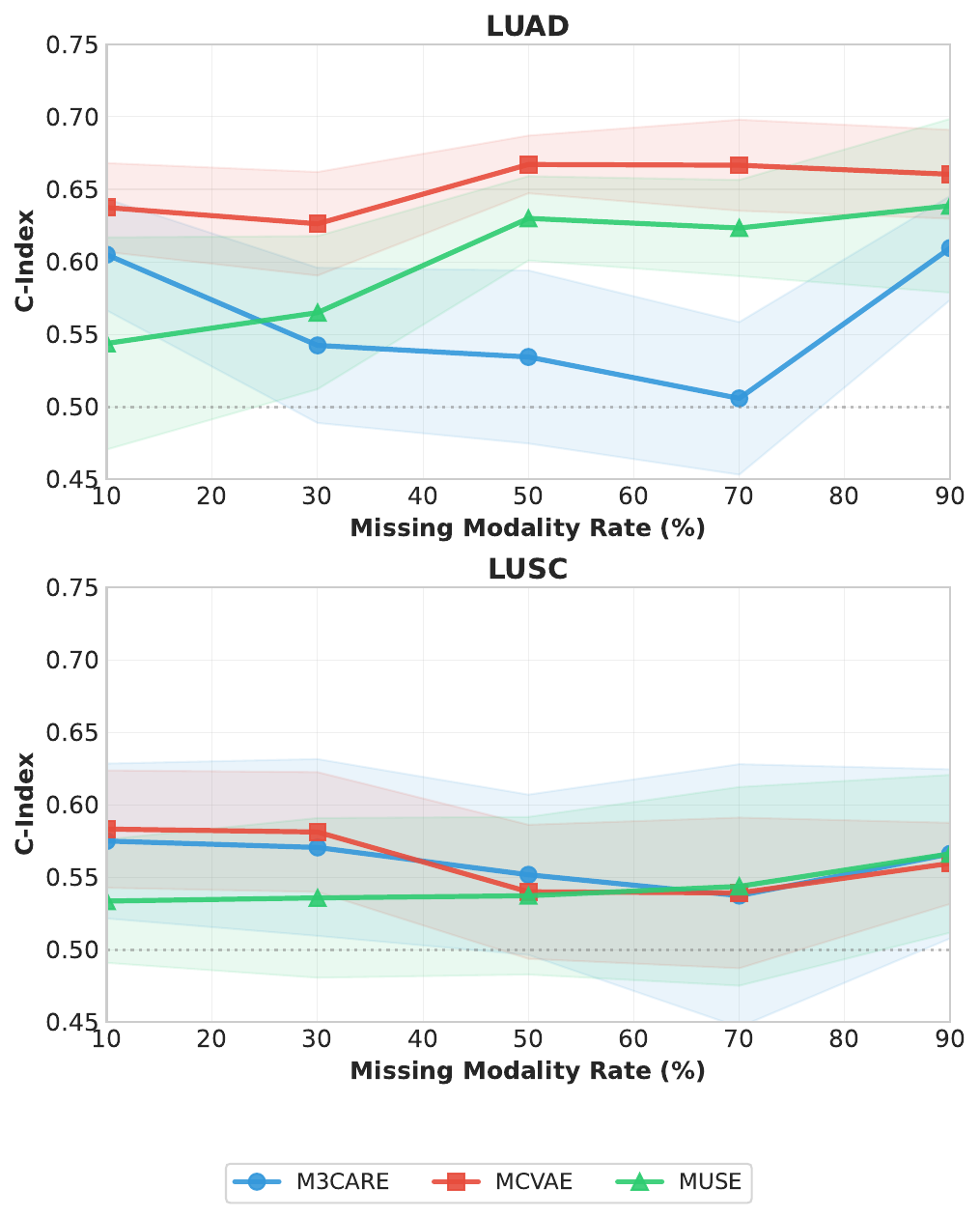}}
\caption{Results on survival analysis for LUAD and LUSC cohorts for different missingness rates $p_{miss}$. Average C-index (bold markers) and standard deviations (shaded area) are reported.}
\label{mod_miss}
\end{figure}

\subsection{Impact of Modality Dropout} 
Figure~\ref{mod_drop} shows how varying dropout rates affect model performance. On LUAD, MCVAE improves from $0.658 \pm 0.036$ (no dropout) to $0.673 \pm 0.029$ ($p_{\text{drop}}= 0.9$), while M3Care and MUSE remain relatively stable (M3Care: $0.624 \pm 0.036$ to $0.585 \pm 0.048$; MUSE: $0.558 \pm 0.078$ to $0.585 \pm 0.073$). 
On LUSC, models diverge slightly at high dropout rates, although MCVAE maintains more stable trajectories than MUSE and M3Care. While there is no significant performance drop at higher dropout rates, MCVAE seems to achieve better results on both cohorts, indicating that consistent masking enforces MCVAE to learn modality-agnostic features that are predictive of the patient's survival.

\subsection{Robustness to Simulated Missingness} 
Figure~\ref{mod_miss} reports model stability under progressively severe missingness during training and testing. Contrary to our expectations, we do not observe a monotonic degrading performance. Overall, MCVAE demonstrates again stable performances even under extreme missingness. In the LUAD cohort, MCVAE consistently achieves a better average C-index across all levels ($0.637\pm0.031$ to $0.660\pm0.031$) than MUSE and M3Care, with a maximum at $50\%$ and $70\%$ missingness ($0.667$), exceeding the performance in the complete-data condition.
M3Care exhibits higher variability in LUAD, indicating that its imputation strategy is less effective at leveraging complementary information under moderate missingness. 
The performance improvement at the highest missingness rate ($90\%$) may indicate that the fusion mechanism in all models compensates for missing modalities by learning from the most available source, i.e., clinical data. 
On LUSC, all models exhibit less variability and tend to converge to similar performance at high missingness ($\approx 0.56$), reflecting task-specific difficulty.
Considering that all models achieve acceptable results with clinical data alone, it may suggest that the impact of missing modalities can thus be mitigated.

\section{Discussion and Limitations}
\subsection{Discussion}
\label{sec:discussion}
In this work, we addressed the critical challenge of missing modalities in clinical settings. This pervasive issue is a result of several factors, including different data acquisition protocols (e.g., sequencing panels for genomic data), privacy concerns, censoring, or limited access to the healthcare system~\cite{Zhang2022}. To systematically evaluate model robustness, we simulate progressively severe missingness patterns that reflect these real-world constraints. Furthermore, we study how different combinations of modalities affect the model's performance, as it is often assumed that integrating more modalities yields better results.

We propose a Multimodal Contrastive Variational AutoEncoder (MCVAE), which leverages a probabilistic framework to explicitly model uncertainty arising from incomplete multimodal observations. The components of our architecture are not new per se, but to our knowledge, they have received limited attention, despite the overall framework being relevant in clinical settings, as it allows for interpretable confidence estimates worth exploring in future studies. 

We evaluate MCVAE against state-of-the-art multimodal learning methods on two non-small cell lung cancer (NSCLC) cohorts. Our results demonstrate consistent improvements across both subtypes, with MCVAE achieving $7.1\%$ and $3.5\%$ higher C-index than MUSE, and $6.8\%$ and $3.1\%$ higher than M3Care, for LUAD and LUSC, respectively. The differential performance between cohorts aligns with recent findings~\cite{Gross2024} that LUAD exhibits more predictable survival patterns than LUSC. These improvements validate that explicit uncertainty modeling and adaptive fusion (rather than static dropout or imputation strategies) yield more robust survival predictions under conditions of missing modality.

When integrating all modalities, MCVAE outperforms both baseline models (Tables~\ref{luad_ablation} and~\ref{lusc_ablation}), though it underperforms the clinical-only baseline on the LUAD cohort. Specifically, the histopathology modality exhibits weak predictive signals and degrades overall performance when integrated into the model. This pattern suggests modality collapse~\cite{Chaudhuri2025}, where conflicting gradient objectives between modalities lead to unbalanced representations that overweight dominant modalities while suppressing weaker but potentially informative signals. Addressing this optimization challenge through improved gradient harmonization or modality-specific learning dynamics is a necessary direction for future work.

We found that increasing modality dropout has a beneficial effect on our model's performance for LUAD, but a marginal effect on LUSC, reflecting the challenges in modeling survival for this histological type. Nonetheless, we notice that under severe missingness conditions, our model has a relatively stable performance, with overall lower variability than the two baselines as shown in Figure~\ref{mod_miss}. We can credit this robustness to several factors: the uncertainty captured by the variational encoders, the fusion bottleneck, which weights each modality based on the learned importance, and, thanks to the contrastive learning objective, to learn modality-invariant features while preserving modality-specific information (e.g., Equation~\ref{eq:contrastive-loss}). Moreover, since we control the missingness rates of all modalities except clinical data, we observe that the performance of our model is close to the unimodal baseline at $90\%$ missingness, which suggests that all models implicitly align with the anchor modality when other sources are unavailable.

\subsection{Limitations and Future Directions}
Our analyses are limited by several factors that warrant consideration. First, our evaluation compared MCVAE with two methods (MUSE and M3Care) and excluded unimodal baselines for histopathology, transcriptomics, and methylation, as these may exhibit different standalone predictive signals. A broader comparison against additional state-of-the-art multimodal methods and systematic unimodal benchmarking would provide a more comprehensive evaluation. Second, our analysis was restricted to NSCLC cohorts from TCGA, a retrospective research database with specific patterns of missingness. Validation across diverse cancer types, and, critically, prospective clinical datasets, is essential for studying clinical applicability.

The benefit of extreme dropout rates deserves deeper investigation. Analyzing learned representations, gate weights, and cross-modal patterns can clarify how extensive masking and reconstruction promote robust feature learning, and in parallel, investigate which modality combinations are most informative for different patient subgroups. The reconstruction component of MCVAE, while serving as an effective regularizer, does not actually impute missing modalities. This design choice avoids potential biases from imputation but also means the model cannot provide estimates of what missing data might look like, which could be valuable for understanding disease patterns. Future work could explore controlled imputation strategies that quantify uncertainty in generated modalities, an approach already established in~\cite{Ma2021}.

An essential evaluation step is represented by the explainability beyond the mechanistic interpretability of our model. Since we aim to discover cross-modal relationships across the studied modalities, we consider applying probing strategies via cross-attention transformers, generalized to multiple modalities. Furthermore, prospective validation of the predictions is necessary to validate the applicability of our model. This includes assessing whether uncertainty estimates align with clinical intuition.

\section{Conclusion}
We presented MCVAE, a framework for multimodal survival prediction that maintains robust performance under simulated incompleteness in multimodal data. On TCGA LUAD and LUSC cohorts, our experiments reveal that multimodal integration does not universally improve performance: in some cases, unimodal or bimodal combinations outperform models trained on all modalities. This opens several avenues for research that focus on the interpretability of multimodal learning strategies dealing with missing modalities and investigate the reasons for such differential performances. In conclusion, MCVAE represents a significant step toward practical multimodal learning for clinical applications, demonstrating robust performance with incomplete data.

\bibliographystyle{IEEEtran}
\bibliography{references}

\appendices

\section{Additional Details on the Implementations}
\label{appendix_implementation}

\subsection{Encoder Architecture}

For modality $k$ with input dimension $d_k$, the encoding process follows:
\begin{equation}
\begin{aligned}
\mathbf{h}_m^{(0)} &= \text{Dropout}_{p}(\mathbf{x}_k), \quad \mathbf{x}_k \in \mathbb{R}^{d_k}, \\
\mathbf{h}_k^{(l)} &= \text{Dropout}_{p}(\text{ReLU}(\text{BN}(\mathbf{h}_k^{(l-1)} ))), \\
\mathbf{z}_k &= \mathbf{h}_k^{(L_k-1)} \in \mathbb{R}^{d_{\text{out}}},
\end{aligned}
\end{equation}
where $\text{BN}(\cdot)$ denotes batch normalization. Dropout rate $p$ and output dimension $d_{\text{out}}$ (fixed for all modalities) are determined through hyperparameter search. Encoders use $L_m = 2$ layers for clinical and WSI features, and $L_m = 3$ layers for transcriptomics and methylation data. Missing modalities are represented as zero embeddings: $\mathbf{z}_m = \mathbf{0} \in \mathbb{R}^{d_{\text{out}}}$.

\subsection{Fusion Network of MCVAE}
The fusion network $h(\cdot)$ in Equation~\ref{eq:adaptive-fusion} transforms modality-specific representations before aggregation. We implement $h(\cdot)$ as a two-layer feed-forward network with residual connections, formally:
\begin{equation}
\begin{aligned}
\mathbf{h}(z^{(k)}) &= z^{(k)} + \text{FFN}(z^{(k)}), \\
\text{FFN}(z) &= \text{Dropout}(\text{GELU}(\text{LN}(z))),
\end{aligned}
\end{equation}
where $\text{LN}(\cdot)$ denotes layer normalization and $d_z$ is the latent dimension. 

\subsection{Decoder Architecture of MCVAE}
Each decoder $g_{\psi_k}: \mathbb{R}^{d_z} \to \mathbb{R}^{d_k}$ reconstruct a masked modality $k$ and consists of $L$ fully-connected layers:

\begin{equation}
\begin{aligned}
\mathbf{h}_k^{(0)} &= z_{\text{fused}}, \\
\mathbf{h}_k^{(l)} &= \text{Dropout}_{p}(\text{ReLU}(\text{BN}(\mathbf{h}_k^{(l-1)}))), \\
\tilde{x}_k^{(k)} &= \mathbf{h}_k^{(L-1)},
\end{aligned}
\end{equation}
where we set $L = 2$ for all modalities, except for clinical features that are not reconstructed.

\section{Additional Details on the Hyperparameters}
\label{appendix_hp}
The tables in this appendix present the values of the hyperparameters selected after a Bayesian search for each model and cohort. All models were trained for a maximum of $150$ epochs with early stopping based on validation C-index and a fixed modality-dropout rate ($p_{\text{drop}} = 0.3$).

\begin{center}
\centering
\noindent\textbf{Table A.1:} Optimized hyperparameters for MUSE.
\label{tab:muse_hp}
\begin{tabular}{lcc}
\toprule
Hyperparameter & LUAD & LUSC \\
\midrule
Embedding dimension ($d_{\text{out}}$) & 128 & 64 \\
Dropout rate ($p$) & 0.148 & 0.427 \\
GNN layers & 4 & 3 \\
Learning rate & $2.86 \times 10^{-4}$ & $1.47 \times 10^{-3}$ \\
Weight decay & $3.97 \times 10^{-4}$ & $1.88 \times 10^{-4}$ \\
Batch size & 16 & 16 \\
\bottomrule
\end{tabular}
\end{center}
\vspace{0.5em}

\begin{center}
\centering
\noindent\textbf{Table A.2:} Optimized hyperparameters for MCVAE.
\label{tab:mcvae_hp}
\begin{tabular}{lcc}
\toprule
Hyperparameter & LUAD & LUSC \\
\midrule
Embedding dimension ($d_{\text{out}}$) & 128 & 128 \\
Hidden dimension & 256 & 256 \\
Dropout rate ($p$) & 0.521 & 0.158 \\
$\beta_{\text{max}}$ (KL weight) & 1.0 & 0.106 \\
Learning rate & $5.28 \times 10^{-5}$ & $1.95 \times 10^{-4}$ \\
Weight decay & $1.24 \times 10^{-4}$ & $5.88 \times 10^{-4}$ \\
Batch size & 16 & 64 \\
\bottomrule
\end{tabular}
\end{center}
\vspace{0.5em}

\begin{center}
\centering
\noindent\textbf{Table A.3:} Optimized hyperparameters for M3Care.
\label{tab:M3Care_hp}
\begin{tabular}{lcc}
\toprule
Hyperparameter & LUAD & LUSC \\
\midrule
Embedding dimension ($d_{\text{out}}$) & 128 & 64 \\
Hidden dimension & 64 & 256 \\
Dropout rate ($p$) & 0.445 & 0.438 \\
GCN layers & 3 & 3 \\
Similarity threshold & 2.056 & 2.380 \\
Learning rate & $2.18 \times 10^{-3}$ & $5.35 \times 10^{-4}$ \\
Weight decay & $4.79 \times 10^{-6}$ & $4.75 \times 10^{-5}$ \\
Batch size & 64 & 16 \\
\bottomrule
\end{tabular}
\end{center}

\end{document}